\documentclass{amsart}

\usepackage{amsmath}
\usepackage{amsthm}
\usepackage{amsfonts}
\usepackage{amssymb}
\usepackage{epsfig}
\usepackage{graphicx} 
\usepackage{palatino,url}

\renewcommand\Re{\operatorname{Re}}

\theoremstyle{plain}

\theoremstyle{remark}

\theoremstyle{definition}

\include{psfig}

\DeclareMathOperator*{\argmax}{argmax}

\begin{document}
\title{Registration of Standardized Histological Images in Feature Space} 


\author[Ba\u{g}c\i]{Ula\c{s} Ba\u{g}c\i}
\address{Collaborative Medical Image Analysis on Grid (CMIAG),The University of Nottingham, Nottingham, UK}
\email{ulasbagci@ieee.org}
\urladdr{www.ulasbagci.net}

\author[Bai]{Li Bai}

\begin{abstract}
In this paper, we propose three novel and important methods for the registration of
histological images for 3D reconstruction. First, possible intensity 
variations and nonstandardness in images
 are  corrected by an intensity standardization process which maps
the image scale into a standard scale where the similar intensities correspond to similar tissues meaning. 
Second, 2D histological images are mapped into a feature space where
continuous variables are used as high confidence image features for accurate registration.
Third, we propose an automatic best reference slice selection algorithm that improves
reconstruction quality based on both image entropy and mean square error of the
registration process. We demonstrate that the choice of reference slice has a significant
impact on registration error, standardization, feature space and entropy information. After
2D histological slices are registered through an affine transformation with respect to an
automatically chosen reference, the 3D volume is reconstructed by co-registering 2D
slices elastically.
\end{abstract}

\maketitle 
\section{INTRODUCTION}
\label{sec:intro}  
2D imaging methods, such as optical microscopy, are still preferable to 3D imaging methods due to their high level of specificity and high resolution properties. Histological sections (slices) obtained through 2D imaging methods provide useful information for the diagnosis or the study of pathology. 3D volume reconstruction from these 2D slices is required in order to fully appreciate anatomical structures~\cite{ourselin00}.

\begin{figure}[h]
   \begin{center}
   \begin{tabular}{c}
\includegraphics[height=5cm]{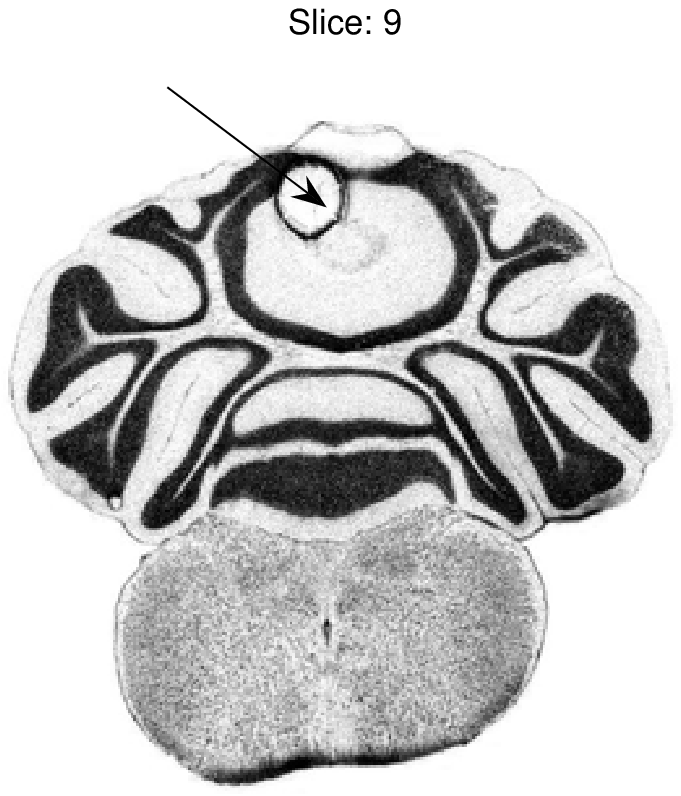} 
\includegraphics[height=5cm]{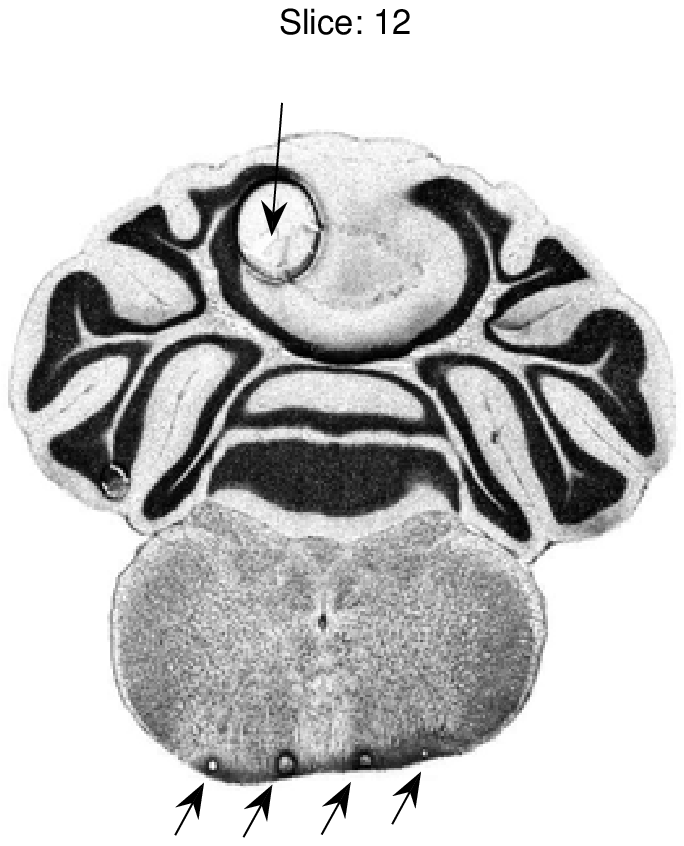}\\ 
\includegraphics[height=5cm]{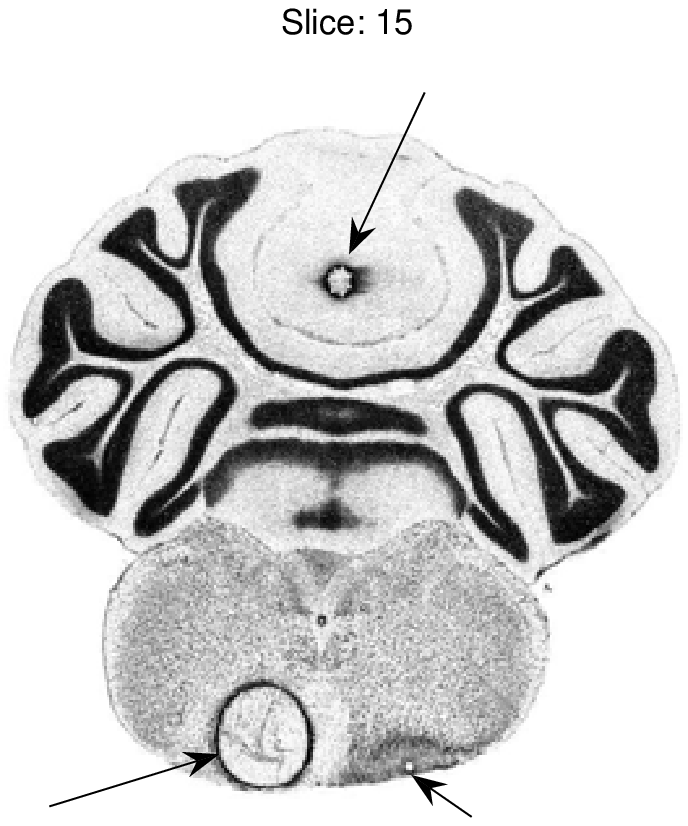} 
\includegraphics[height=5cm]{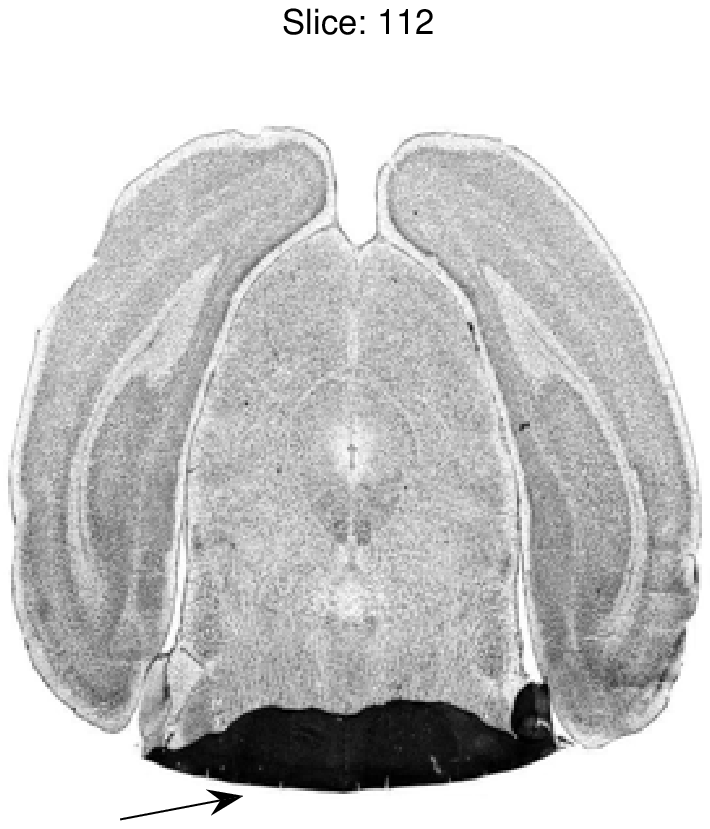}\\
   \end{tabular}
   \end{center}
\caption{Histological images with non-cohorent distortions, tears, hole and missing parts \label{img:badslices}}
\end{figure}

Typically, a 3D volume is reconstructed by registering (aligning) the 2D sections with respect to a chosen reference and stacking successive aligned sections~\cite{malandain04}. As the acquisition processes of different 2D histological images are performed independently, slice misalignment and deformation is often unavoidable. Figure~\ref{img:badslices} shows examples of histological slices with non-cohorent distortions, tears, hole and missing parts. The deformation varies from section to section and non-cohorent distortions may exist in consecutive sections. Choosing an arbitrary slice as a reference slice leads to errors in 3D volume reconstruction, hence,  the reference slice should be chosen properly not to contain distortions in order to achive high quality volume reconstruction~\cite{bagci_report}.

Automatic registration of histological slices are necessary because manual registration using interactive alignment is non-reproducible and user dependent, therefore, it cannot be used if the number of sections is large~\cite{malandain04,ourselin00}. Among various automated registration methods proposed in the literature, rigid and affine registration methods can only handle global deformations and the transformation recovered from rigid registration has no clinical significance. Since histological slices change smoothly from slice to slice and the section distortions induced by the preparation process are local in nature~\cite{ju06, bagci_report}, accurate alignment of these slices can be achived by using elastic registration methods~\cite{gee97,elastic1,nonmedsurvey,periaswamy,superfast}.

\begin{figure}[h]
   \begin{center}
   \begin{tabular}{c}
\includegraphics[height=5cm]{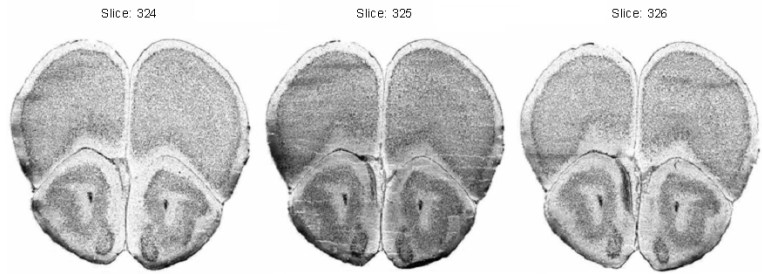}
   \end{tabular}
   \end{center}
\caption{Consecutive slices with different contrast map even for the same tissue level \label{img:contrastdiff}}
\end{figure}

Although intensity differences in consecutive slices are generally assumed to be small and are often ignored in most of the existing nonrigid registration methods, it is shown that intensity variations for the same tissues can lead to large registration errors and the reconstructed 3D volume will not be smooth~\cite{bagci_report, bagci07,elastic1,nonmedsurvey,ju06, malandain04,ourselin99}. Figure~\ref{img:contrastdiff} demonstrates that consecutive slices may have different brightness/contrast characteristics even for the same tissue regions. Similar to our previous study on MRI~\cite{bagci07}, to overcome intensity variations, we perform registration of histological images in standard intensity scale where similar intensities represent similar tissues.

To ensure that reconstructed 3D volume represents the full anatomy, one way is to superimpose the reconstructed volume onto an unsectioned reference volume, if it exists~\cite{malandain04}. However, the problem with this approach is that a reference volume is not always available~\cite{ju06}. In the absence of a reference volume, qualitative evaluation of reconstructed volume is error prone and quantitative evaluation is often needed. The smoothness of the reconstructed surface/volume can be used as a quantitative evaluation method in that case~\cite{ju06}. Therefore, we use smoothness based measurement method, Correlation Alignment Measure (CAM), to quantify reconstructed volume.

A novel 3D reconstruction method for histological images is described in this paper. The method tackles three difficult problems in registration of histological images. First, in order to capture intensity variations between slices, we standardize histological images (Section~\ref{sec:std}). Second, classical motion estimation based affine and locally affine registration methods are used to register images in feature space and image space respectively (Section~\ref{sec:feat}). Third, we propose an automatic best reference slice (BRS) selection algorithm based on iterative assessment of image entropy and mean squre error (MSE) of the registration process in order to improve the quality of the reconstructed volume (Section~\ref{sec:bestref}). Quantitative evaluation of reconstructed volume is given in Section~\ref{sec:results}. The paper is concluded with future research directions and conclusion (Section~\ref{sec:conc}).

\section{Standardization of Image Intensity Scale}
\label{sec:std}
Image intensity variations are not only influenced by the distribution of light sources, but also the content (different tissues) of the images as different tissues show different intensity levels. It is important to transform the image scale into the standard intensity scale so that for the same body region, intensities will be similar~\cite{bagci_report, bagci07, udupa_std_jmri}. This is called the intensity standardization. As can be seen from Figure~\ref{img:contrastdiff}, some edges are clearly visible in one image, but are not visible in the other. This shows the need for intensity standardization.

Standardization is a non-linear pre-processing technique which maps image intensity histogram (scale) into a standard intensity histogram through a training and a transformation step. In the training step, a set of images of the same body region are given as input to "learn" histogram-specific parameters, or the landmarks. In the transformation step, any given image is standardized with the estimated histogram-specific landmarks obtained from the training step.

It has been proven that image histograms of the same body region are always of the same type, and most of the histograms of biomedical images are bimodal~\cite{udupa_std}. Since the histological rat brain images we study also produce bimodal intensity histograms, we extract histogram specific landmarks according to bimodal distribution as suggested in~\cite{udupa_std_jmri, udupa_std}. 

\begin{figure}[h]
 \begin{center}
   \begin{tabular}{c}
\includegraphics[height=5cm]{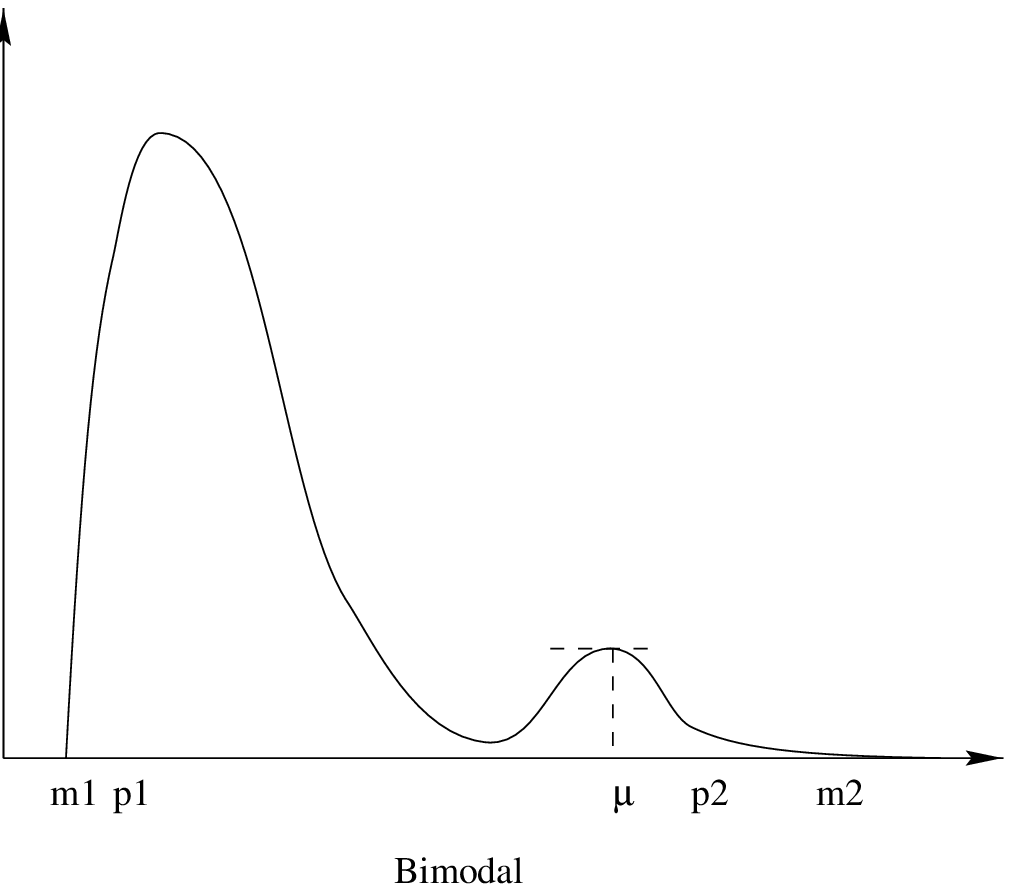} 
\includegraphics[height=5cm]{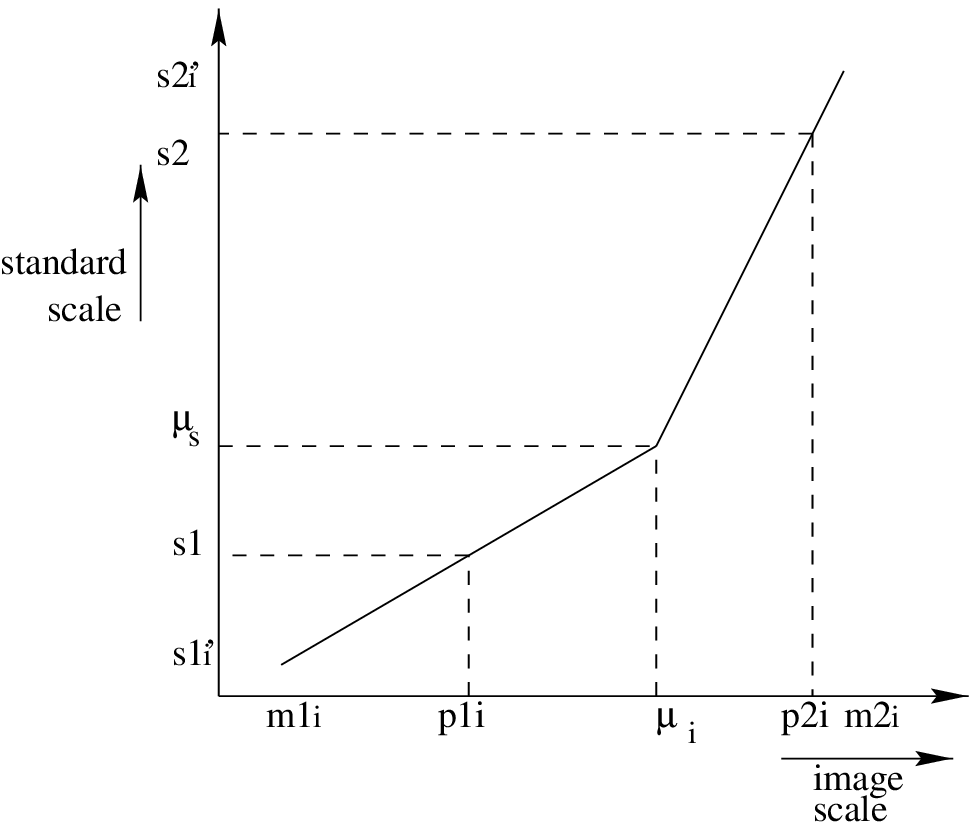} 
   \end{tabular}
   \end{center}
\caption{a) Location of the histogram specific landmarks, $m_1$=minimum gray value, $m_2$=maximum gray value.  b) The intensity mapping function for the transformation step \label{img:bimodal}}
\end{figure}

In bimodal histograms, one of the histogram specific landmarks is the mode ($\mu$) representing the main foreground object in the image, as depicted in Figure~\ref{img:bimodal}(a). 
Other histogram specific landmarks denoted by $p_1$ and $p_2$ are extracted according to the range of intensity of interest (IOI) by setting minimum and maximum percentiles values, namely $pc_1$ and $pc_2$. For any image $\mathbf{F}^{(j)}$, we will consider histogram specific landmarks as:
\begin{equation} 
L_j = \left\lbrace p_{1j}, p_{2j}, \mu_j \right\rbrace, \quad j=1,..,M
\end{equation}
where we can increase the number of histogram specific landmarks by setting more percentiles values for the main foreground $\mu$~\cite{udupa_std_jmri,udupa_std}.

In the training step, for each image, the landmarks $L_j = \left\lbrace p_{1j}, p_{2j}, \mu_j \right\rbrace$ obtained from the histogram $H^{(j)}$ are mapped into the standard scale by mapping intensities from $[p_{1j}, p_{2j}]$ to $[s_1, s_2]$ where $s_1$ and $s_2$ are minimum and maximum intensities on the standard scale respectively. The formula for mapping $x\in [p_{1j}, p_{2j}]$ to $x'\in [s_1, s_2]$ is the following~\cite{udupa_std_jmri}.
\begin{equation}
\label{eq:mapping}
x'=s_1 + \frac{x-p_{1j}}{p_{2j}-p_{1j}}(s_2-s_1)
\end{equation}
Figure~\ref{img:bimodal}(b) shows the two linear mappings. The first from $[p_{1i},\mu_i]$ to $[s_1,\mu_s]$ and the second from $[\mu_i, p_{2i}]$ to
$[\mu_s, s_2]$. Overall mapping, $\tau_i(x)$, from $[m_{1i}, m_{2i}]$ to $[s_{1i}', s_{2i}']$ can be summarized as follows:
\begin{equation}
\label{eq:finalmapping}
\tau_i(x)= \left\{ \begin{array}{ll}
\lceil \mu_s+ (x-\mu_i)\left( \frac{s_1-\mu_s}{p_{1i}-\mu_i}\right) \rceil &
\textrm{if $m_{1i} \leq x \leq \mu_i$}\\
\lceil \mu_s+ (x-\mu_i)\left( \frac{s_2-\mu_s}{p_{2i}-\mu_i}\right) \rceil &
\textrm{if $\mu_i \leq x \leq m_{2i}$}
\end{array} \right.
\end{equation}
where $\lceil . \rceil$ converts any number y$\in \Re$ to the closest integer Y
such that Y $\geq y$ or $\leq y$. Further details and proofs can be found
in~\cite{udupa_std}.

\begin{figure}[h]
 \begin{center}
   \begin{tabular}{c}
 \includegraphics[height=6cm]{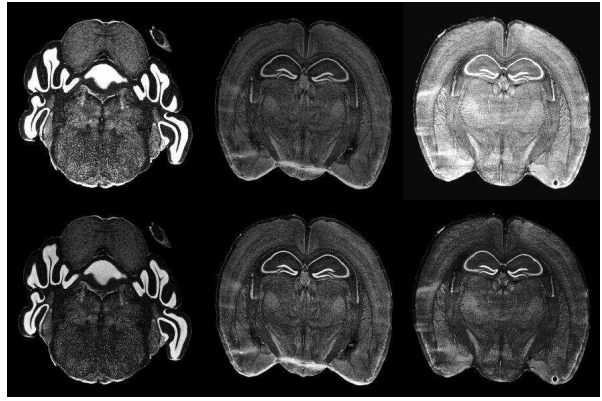}
   \end{tabular}
   \end{center}
\caption{Various Original slices before (first row) and after standardization (second row).\label{img:beforeandafter}}
\end{figure}

Figure~\ref{img:beforeandafter} shows slices before and after standardization. The first row shows the original data displayed using the default window setting. The second row shows the same slices after standardization displayed using the "standard" window settings with the parameters defined in Section~\ref{sec:results}.

\section{Registration in Feature Space}
\label{sec:feat}
Registration of histological slices requires serial registration procedure which is a combination of transformations.
Let $A_{j \leftarrow i}$ be the transformation that warps the source image $i$ to the target image $j$. The transformation $A_{j \leftarrow i}$ is computed serially as follows:
\begin{eqnarray}
A_{j \leftarrow i} = A_{j \leftarrow j-1} \circ  A_{j-1 \leftarrow j-2} \circ \ldots A_{i+1 \leftarrow i},\qquad if  \quad i<j\nonumber \\
A_{j \leftarrow i} = A_{j \leftarrow j+1} \circ  A_{j+1 \leftarrow j+2} \circ \ldots A_{i-1 \leftarrow i},\qquad if \quad  i>j
\end{eqnarray}
where $\circ$ represents composition.

\subsection{Locally Affine Nonlinear Transformation}
Since consecutive slices are not exactly the same, rather slices vary smoothly, locally affine globally smooth (LAGS), registration algorithm fits well to the problem. LAGS\footnote{also known as elastic registration} registration algorithm uses 8-affine parameters to fully represents changes between 2D images. Two of these affine parameters are needed to capture local brightness and contrast patterns~\cite{periaswamy} and six affine parameters are used to capture local deformations for 2D images. As described in our previous study~\cite{bagci07}, there is no need to use these 2 affine parameters because the standardization procedure is used to remove intensity variations among the same tissue types. Readers are encouraged to read~\cite{periaswamy,bagci07} to understand theory of LAGS and the modified algorithm which takes into account the standardization procedure. 

Choice of the feature space plays a vital role in image registration especially if the similarity metric bases on the optimization function independent of spatial information (i.e mutual information). Since these kind of registration methods do not take into consideration the spatial information of pixel/voxel intensity distribution/variation, the optimization algorithm may get stuck in local maximum resulting in misalignment. Defining a feature space capturing variations of gray-level characteristics will overcome the drawbacks of intensity based approaches. To align the images globally, we used a particular feature space which represents an image by continuous variables, called edgeness, and describes the intensity variance of a predefined region over the image~\cite{thiran01, bagci_report}.

\subsection{Notation and Formulation}
\begin{figure}[h]
 \begin{center}
   \begin{tabular}{c}

\includegraphics[height=5cm]{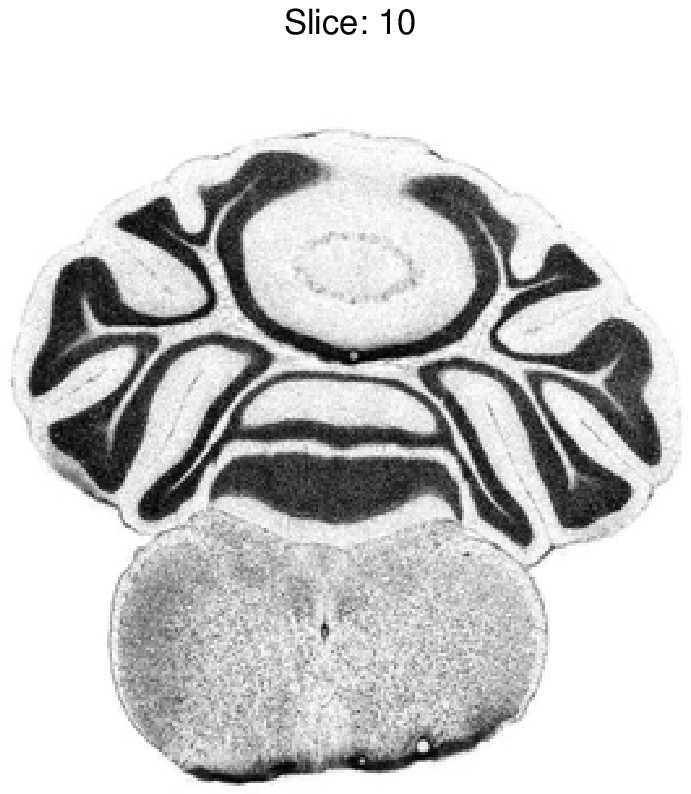} 
\includegraphics[height=5cm]{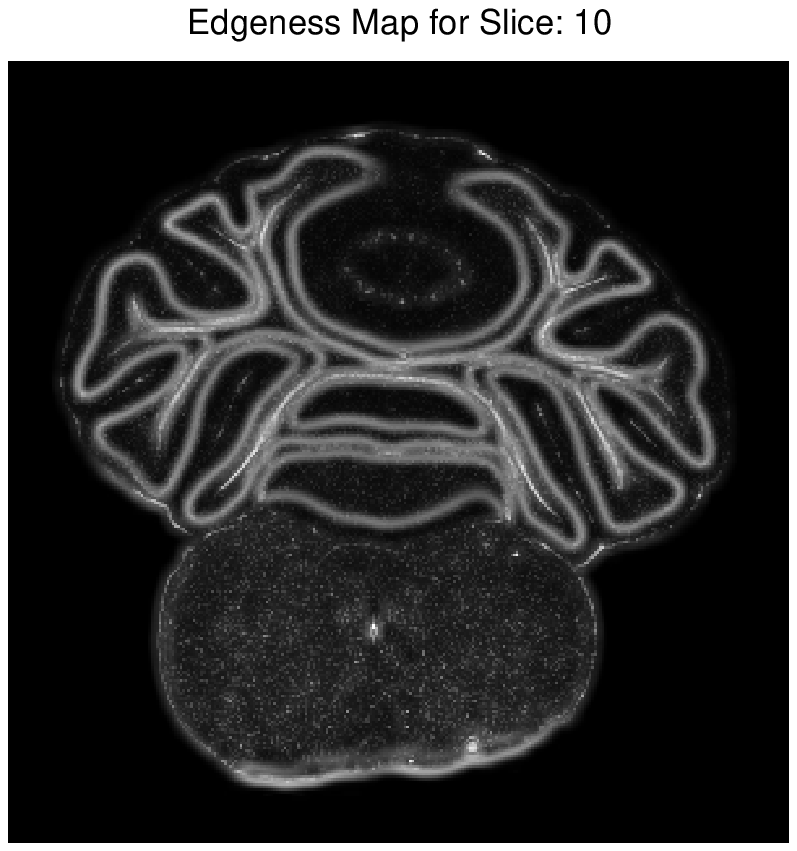}\\ 
\includegraphics[height=5cm]{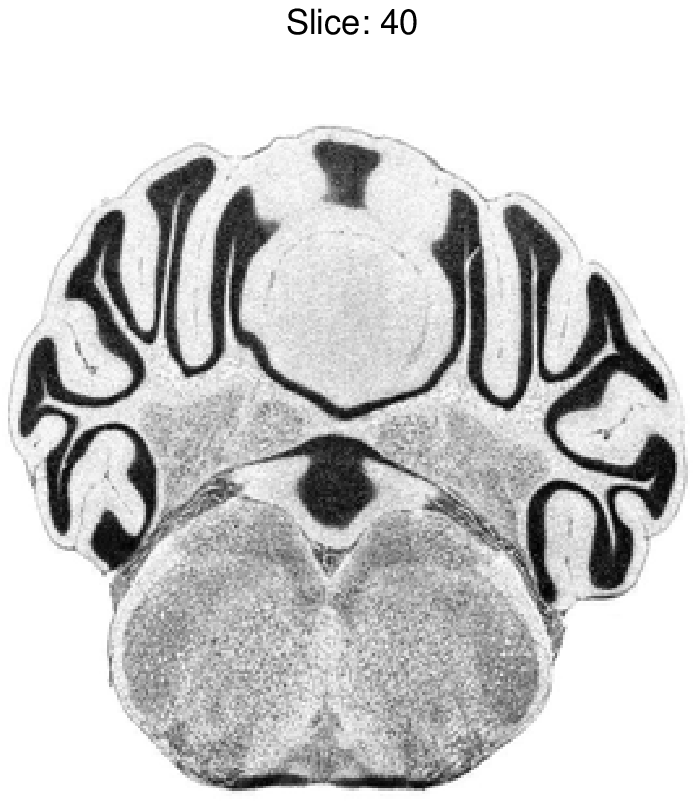} 
\includegraphics[height=5cm]{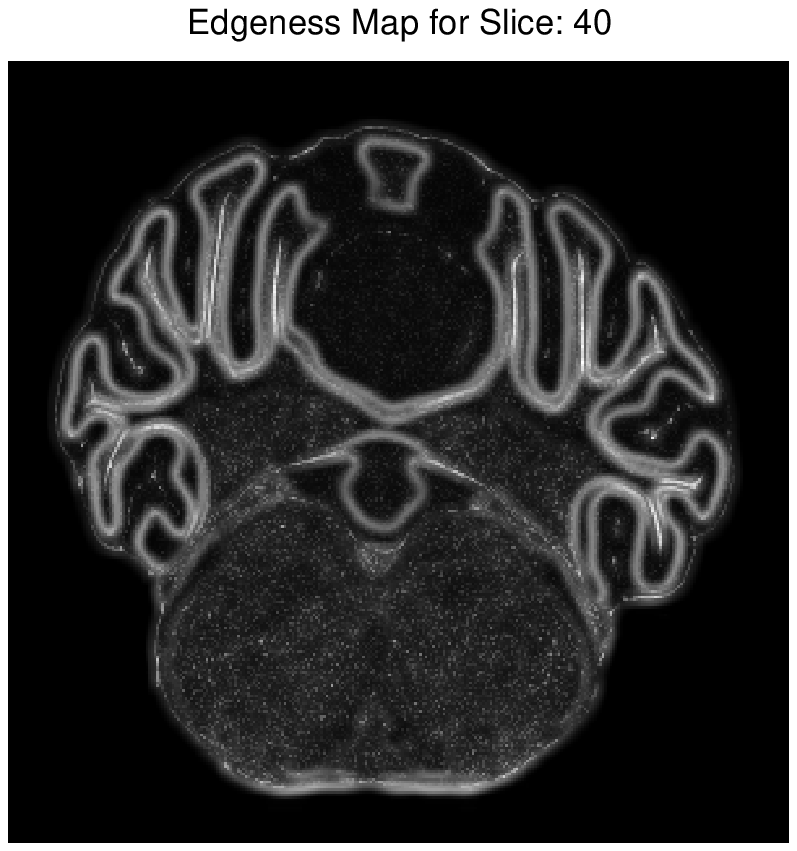}\\
   \end{tabular}
   \end{center}
\caption{Histological images and their edgeness maps with radius $|r_f|$=3 pixels\label{img:edgeness}}
\end{figure}

We represent an image (section/slice) by a pair $\mathbf{F}=(F,g)$ where $F$ is a two-dimensional (2-D) array of pixels and  $g$ is intensity function defined on $F$, assigning an integer intensity value for each pixel $o \in \mathbf{F}$. A particular feature space that allows the representation of an image by continuous variables is called the edgeness space~\cite{thiran01}, which describes the intensity variance of local regions in an image. At image coordinate $r_0$, the edgeness is represented by
\begin{equation}
\label{eq:edgeness}
\mathbf{F_e} = \sum_{|r_i - r_0|<r_f} | g(r_i) - g(r_0)|,
\end{equation} 
\noindent where $r_f$ represents a fixed radius. Note that this is not just to determine whether a specific voxel/pixel is edge or not~\cite{thiran01}. Instead, within a specified radius $r_f$, the image feature content is forced to stay beyond a variation level
which prevents the registration process from getting stuck in local maxima. Edgeness maps for various slices are shown in Figure~\ref{img:edgeness}.

\section{Automatic Best Reference Slice Selection}
\label{sec:bestref}
The quality of the reconstructed 3D volume mostly depends on the choice of the reference slice. The reference slice is used as a target image and all the remaining slices are being considered as source images to be registered onto the target image. If the reference slice is distorted or noisy, reconstructed 3D volume will not be optimal. Once the reference slice is identified as target image, registration based fusion methodology can be applied for reconstruction~\cite{malandain04}.  

\subsection{Metrics}
\label{subsec:metrics}
Selecting best reference slice can be based on high confidence image features such as MSE, entropy, edge, texture, color, intensity histograms, etc.
\begin{enumerate}
\item \textbf{MSE:} In the case of distortions, structural discontinuity is not minimum even for the consecutive slices. When affine registration is performed for global alignment of images, the optimization procedure tries to minimize MSE between images but due to distortion, it will not reach low MSE values. Furthermore, it is also known that with small SNR values, alignment is difficult, leading to high registration errors. Therefore, MSE can be used as a tool for checking whether the slices are distorted or not. While high MSE values indicate most probably distorted and noisy slices, low MSE values indicate strong similarity between consecutive images. 

\item \textbf{Edge:} In feature space, we emphasise edgeness
features of an image by mapping image space into the feature space where edgeness parameters hold both edge information and spatial variations of pixel intensities over all regions in the image. Therefore, we assume that MSE between any image pair already includes high confidence information related to edges.

\item \textbf{Contrast/Brightness:} Contrast/brightness patterns also play important role in image contents. Since standardization method has been used to correct intensity variations, intensity for the same tissue is the same for all images. 
 
\item \textbf{Entropy:} Entropy is another measure often used to characterise the information content of a data source. It has been used as a metric for image registration in the form of mutual information. Large mutual information between images implies high similarity and vice versa.
\end{enumerate}
A common way to define entropy $E$ for the $jth$ image $\mathbf{F}^{(j)}$ is:
\begin{equation}
\label{feat_compute}
E^{(j)} = -\sum_{o\in I}p_o(\mathbf{F}^{(j)})log(p_o(\mathbf{F}^{(j)}))
\end{equation} 
where $p_o(.)$ is the probability density function for intensity value $o$ and $I$ is set of all possible gray values in $\mathbf{F}$. It is certain that if the amount of information in the image is large, the entropy value will be high. From this point, we propose the combination of entropy information with MSE as a high confidence feature measure to automatically identify BRS for 3D volume reconstruction.   

\subsection{Best Reference Slice (BRS)}
\label{subsec:BRS}
To find the BRS according to the feature sets defined in Section~\ref{subsec:metrics}, an iterative framework is developed to compare slices that are not consecutive and to avoid
errors arising from noisy and distorted images. The framework divides slices into subvolumes. Let $M$ be total number of slices, $V_{k}$ the $kth$
subvolumes where $k=1,..,vol$ and $N_k$ the number of slices in subvolume $k$ such that: $ \sum_{k=1}^{vol}N_k=M $.

Since MSE is inversely related to the similarity of images and entropy information is directly related to the content of the image, we define:
\begin{equation}
BRS_k = \argmax_{i\neq j\in V_k} \left\lbrace log\left(
\frac{E^{(j)}}{MSE_{i,j}}\right) \right\rbrace
\end{equation}
where $MSE_{i,j}$ is feature space based mean square error after registering the image $i$ with image $j$ and
\begin{equation}
\label{eq:maxentropy}
j=\argmax_{s \in V_k}\left\lbrace E^{(s)}\right\rbrace
\end{equation}
for subvolume $V_k$. 

\section{Experiments and Results}
\label{sec:results}
\subsection{Rat Brain Data}
We have registered a stack of 350 Nissl-stained slices acquired by cyro-sectioning coronally an adult mouse brain with a resolution of 590x520 pixels at a resolution of 15$\mu$m and 24-bit color format \cite{ratatlas}.

\subsection{Standardization Parameters}
Based on the experiments in~\cite{udupa_std_jmri, udupa_std}, minimum and maximum percentile values are set to $pc_{1}=0$ and $pc_2=99.8$ respectively. In the standard scale, $s_1$ and $s_2$ are set to $s_1=1$ and $s_2=4095$. Examples of intensity mapping for a couple of images, source and target, can be found in~\cite{bagci07} where it can be seen that after standardization, histograms are more similar in shape and location. It means that intensities have tissue-specific meaning after standardization~\cite{bagci_report, bagci07, udupa_std_jmri, udupa_std}. 

\subsection{Implementation}
Briefly, registration is performed initially for slices in each subvolume separately (See Section~\ref{subsec:BRS}). Three kinds of registration are performed in the reconstruction process: rigid, affine and LAGS. MSEs are calculated according to affine registration in edgeness space and is used to select BRSs for each subvolume. Affine registration is performed in a serial manner combining transformation functions.  Then, LAGS registration is performed to capture local deformations in each subvolume with respect to the chosen reference. Once LAGS registration processes are completed, subvolumes are registered to each other in a rigid manner. 

In order to achieve low computational cost and accelerate the registration process, coarse-to-fine multiresolution framework is used. Registration in finer level is performed with the result of the previous level as initial condition. This process continues until the finest level is reached. To ensure a more accurate solution, we perform standardization after each warping/interpolation. Either small or large, intensity changes caused by the interpolation are captured by standardization~\cite{bagci07}.

\subsection{Evaluations}
\label{subsec:eval}
Quantitative evaluation of the results of the reconstruction process is often difficult. It has been shown in~\cite{guest01} that an ideal measure of the quality of the reconstruction is the smoothness of the reconstructed surfaces. In that work, they propose a new measure based on evaluation of smoothness of the reconstructed volume called \textit{Correspondence Alignment Measure} (CAM).

The CAM measure relies on the assumption that if a point is perfectly aligned, it lies midway between its corresponding points on neighbors' sections. To compute the CAM measure for a given image, first of all, corresponding points for specifed control points in the image are identified. The associated confidence values in two adjacent images are then calculated. If the confidence is  greater than a pre-defined threshold $\tau$, square root of the summation of the deformation vectors are added to the cumulative sum. Finally, the cumulative sum is normalized by the number of pixels which have contributed.  Note that CAM gives one value for each image, therefore, mean or standard deviation of CAM values of serial images are needed to compare reconstructions. Reconstucted volume is smooth if the mean or the standard deviation of CAM measures are low and vice versa. 

Summary of the changes in mean and standard deviation in CAM values is given by Table~\ref{table:cam}. The values in Table~\ref{table:cam} are obtained by considering the worst case which uses all the slices instead of just a few slices from the middle of the stack as defined in~\cite{guest01}, and $\tau$ is set to $0$. Even for the worst case, CAM values indicate that a smooth volume is constructed with the proposed framework. While mean values dropped by $7.29$\% and $18.69$\%, the standard deviation values dropped by $24.46$\% and $27.73$\% for affine and locally affine registered stacks respectively with respect to rigid registered stacks. Figure~\ref{img:cam} shows CAM values for each section. Registered stacks has lower CAM values which means that smoother reconstructed surface/volume is obtained.

\begin{figure}[h]
\label{img:cam}
 \begin{center}
   \begin{tabular}{c}
\includegraphics[height=6cm]{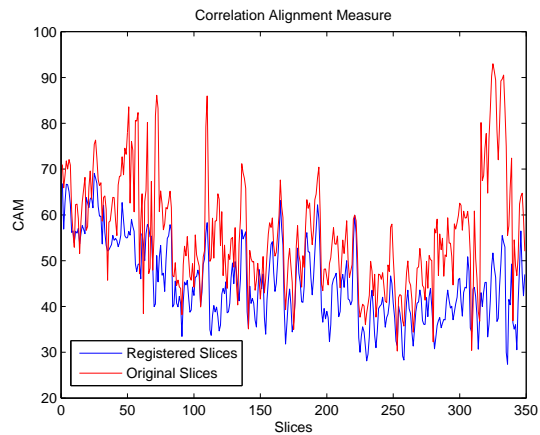} 
    \end{tabular}
	\end{center}
\caption{CAM for original and registered slices\label{img:cam}}
\end{figure}

\begin{table}[t]

\caption{CAM-mean and standard deviation values for the reconstructed 3D\label{table:cam}}
\begin{center}

\begin{tabular}{|c|c|c|c|}
\hline
-- & Rigid Reg.& Affine Reg. & LAGS Reg. \\ \hline 
Mean                    &  55.911         & 51.832  & 45.461 \\  \hline
Std                       &  12.223         &  9.232    & 8.833  \\ \hline
\end{tabular}
\end{center}
\end{table}

\section{Conclusion}
\label{sec:conc}
In this study, we present a novel method for reconstructing 3D rat brain volumes from 2D histological images. The framework bases on three fundamental premises. (1) All histological images must be standardized for accurate registration leading to 3D volume reconstruction. (2) For accurate and succesful registrations in consecutive slices, a reliable feature space must be taken into account. (3) For automatic 3D volume reconstruction, the reference slice must be chosen properly by avoiding slices with high noise, distortions and other factors. To validate the reconstructed volume, the smoothness of the volume is considered.  Experimental results indicate that the reconstructed volume is highly accurate.

\section{Acknowledgements}
This research is funded by the European Commission Fp6 Marie Curie Action Programme (MEST-CT-2005-021170) under the CMIAG (Collaborative Medical Image Analysis on Grid) project.

We wish to thank Prof. Jayaram K. Udupa, Image Processing Group of University of Pennsylvania, for his comments and suggestions.



\end{document}